%% file: main.tex
\documentclass[10pt, conference]{IEEEtran}

\usepackage{cite}
\usepackage{amsmath,amssymb,amsfonts}
\usepackage{algorithmic}
\usepackage{graphicx}
\usepackage{textcomp}
\usepackage{pdfpages}
\usepackage{multirow}
\usepackage{subcaption}
\usepackage{booktabs}
\usepackage{epsfig}
\usepackage{gensymb}
\usepackage{tabu}
\usepackage{pgfplots}
\usepackage{caption}
\usepackage{algorithm2e}
\usepackage{url}

\usepackage[acronym,nowarn]{glossaries}
\makeglossaries
\loadglsentries{glossary}
\pgfplotsset{compat=1.18}

\def\BibTeX{{\rm B\kern-.05em{\sc i\kern-.025em b}\kern-.08em
    T\kern-.1667em\lower.7ex\hbox{E}\kern-.125emX}}
\begin{document}

\title{Classifying Dry Eye Disease Patients from Healthy Controls Using Machine Learning and Metabolomics Data}

\author{\IEEEauthorblockN{Sajad Amouei Sheshkal}
\IEEEauthorblockA{\textit{Department of Computer Science} \\
\textit{Oslo Metropolitan University (OsloMet)}\\
Oslo, Norway \\
sajad.amouei@gmail.com}
\and
\IEEEauthorblockN{Morten Gundersen}
\IEEEauthorblockA{\textit{Department of Life Sciences and Health} \\
\textit{Oslo Metropolitan University}\\
Oslo, Norway \\}
\and
\IEEEauthorblockN{Michael Alexander Riegler}
\IEEEauthorblockA{\textit{Department of Computer Science} \\
\textit{Oslo Metropolitan University (OsloMet)}\\
Oslo, Norway \\}
\and
\IEEEauthorblockN{Øygunn Aass Utheim}
\IEEEauthorblockA{\textit{ Department of Ophthalmology} \\
\textit{Oslo University Hospital}\\
Oslo, Norway \\}
\and
\IEEEauthorblockN{Kjell Gunnar Gundersen}
\IEEEauthorblockA{\textit{Ifocus Eye Clinic} \\
Haugesund, Norway \\}
\and
\IEEEauthorblockN{Hugo Lewi Hammer}
\IEEEauthorblockA{\textit{Department of Computer Science} \\
\textit{Oslo Metropolitan University (OsloMet)}\\
Oslo, Norway \\}
}

\maketitle

\begin{abstract}
Dry eye disease is a common disorder of the ocular surface, leading patients to seek eye care. Clinical signs and symptoms are currently used to diagnose dry eye disease. Metabolomics, a method for analyzing biological systems, has been found helpful in identifying distinct metabolites in patients and in detecting metabolic profiles that may indicate dry eye disease at early stages. In this study, we explored using machine learning and metabolomics information to identify which cataract patients suffered from dry eye disease. As there is no one-size-fits-all machine learning model for metabolomics data, choosing the most suitable model can significantly affect the quality of predictions and subsequent metabolomics analyses. To address this challenge, we conducted a comparative analysis of eight machine learning models on three metabolomics data sets from cataract patients with and without dry eye disease. The models were evaluated and optimized using nested k-fold cross-validation. To assess the performance of these models, we selected a set of suitable evaluation metrics tailored to the data set's challenges. The logistic regression model overall performed the best, achieving the highest area under the curve score of $0.8378$, balanced accuracy of $0.735$, Matthew's correlation coefficient of $0.5147$, an F1-score of $0.8513$, and a specificity of $0.5667$. Additionally, following the logistic regression, the XGBoost and Random Forest models also demonstrated good performance.
\end{abstract}


\section{Introduction}
 
\gls{ded} is a multifaceted disorder characterised by a disruption in the composition, integrity, and stability of the tear film due to various internal and external factors. It is one of the most common reasons people seek eye care, with a severity spectrum ranging from minor, fleeting discomfort to severe, persistent pain and visual function impairment. This progression not only presents a substantial economic and healthcare challenge but also significantly impacts the quality of life of sufferers and the broader community. The incidence of \gls{ded} notably increases following cataract surgery, highlighting the critical need for ophthalmologists to thoroughly evaluate for existing \gls{ded} and to implement proactive treatment approaches. The presence of \gls{ded} before surgery, can also complicate the precision of pre-surgical measurements, necessitate the reduction of intra-operative factors that could harm the ocular surface, and require the adoption of post-surgical care protocols to prevent the worsening of \gls{ded} symptoms\cite{yazdani2019tear,naderi2020cataract,zeev2014diagnosis,dana2019estimated,gomes2019impact,zheng2017prevalence}. Clinical signs and symptoms are currently used to diagnose dry eye disease; however, the correlation between signs and symptoms is weak, leading to challenges in diagnosing and monitoring \gls{ded} \cite{wolffsohn2017tfos}.

Advancements in omics technologies allow researchers to explore the genome, transcriptome, proteome, and more, providing in-depth insights into the molecular mechanisms underlying diseases. Despite their utility, single omics approaches are insufficient for comprehensively understanding the intricate interactions between genes, RNA, proteins, and environmental factors. Metabolomics distinguishes itself by revealing how metabolites, the dynamic output of gene, mRNA, and protein function, respond to various internal and external stimuli. This makes metabolomics a crucial instrument for unraveling complex biological processes and deepening our understanding of the origins, progression, and treatment outcomes of diseases \cite{yazdani2019tear}.

In the context of \gls{ded}, metabolomics holds the promise of identifying disease-specific metabolite profiles. These profiles can play an important role in enhancing the early diagnosis of \gls{ded}, and in elucidating its etiology and pathology \cite{yazdani2019tear}. By identifying specific metabolic pathways and therapeutic targets, metabolomics can guide the choice of personalized treatment plans and improve the prediction of patient outcomes \cite{yazdani2019tear}. Furthermore, this approach can significantly enhance the monitoring of disease progression and the evaluation of treatment efficacy, ultimately leading to more effective management of \gls{ded} \cite{yazdani2019tear,smolenska2015metabolomics}. By comparing metabolomic data from patients with that of a control group, researchers and clinicians can gain valuable insights, further enriching the biomedical and clinical understanding of \gls{ded}.

In metabolomics research, the use of machine learning models is important for unraveling complex metabolic networks and identifying biomarkers. Each machine learning algorithm offers unique capabilities, necessitating careful selection to ensure the effectiveness of a study \cite{galal2022applications}. The choice of a machine learning algorithm significantly impacts the investigation's success, as it must align with the specific characteristics of the data, including its dimensionality and the biological complexity \cite{galal2022applications}. Researchers dedicate efforts to evaluate the predictive performance of various machine learning algorithms, comparing them to traditional statistical methods to identify the most suitable approach for their specific needs \cite{shah2021review}.

However, selecting the optimal machine learning algorithm for metabolomics studies is challenging. Comparative studies often reveal that the effectiveness of a machine learning model can vary greatly depending on the research context and data attributes. This inconsistency underlines the fact that there is no one-size-fits-all machine learning algorithm for metabolomics \cite{mendez2019comparative,delafiori2021covid,yagin2023explainable,hu2022explainable,tiedt2020circulating}. Researchers are encouraged to develop a nuanced understanding of the advantages and operational intricacies of each machine learning model. By tailoring their choice of algorithm to the specific demands and nuances of their data, scientists can enhance the accuracy and interpretability of their findings \cite{galal2022applications}. Despite the lack of a universal guideline for the selection of machine learning algorithms, this iterative process of assessment and application is essential to advance metabolomics research and discover new biological insights.

This study focused on evaluating the efficacy of various machine learning models to classify cataract patients based on the presence or absence of \gls{ded}, utilising metabolomics data collected from individuals awaiting cataract surgery. This approach marks an effort to explore the application of machine learning methodologies to this specific group of patients, to our knowledge, not previously undertaken. 

Identifying effective machine learning models can help more accurately classify cataract patients with and without \gls{ded}. This can also provide more reliable machine learning models for use by explainable artificial intelligence methods in identifying metabolomics signatures.

The following are our main contributions:
\begin{itemize}
    \item The most common machine learning techniques in the metabolomics field have been employed on imbalanced metabolomics data sets to classify cataract patients, differentiating between those who have dry eye disease and those who do not.
    \item Machine learning models have been fine-tuned to achieve better performance and to aim at selecting the best models.
    \item Metabolomics data obtained from positive and negative ionization modes have been evaluated to analyse their impact on dry eye identification performance.
    \item A set of performance metrics, suitable for the data set's challenges, has been selected to appropriately evaluate the developed models.
\end{itemize}

Our research methodology involved an initial preprocessing phase, in which the data sets were standardised and normalised. Subsequently, we employed the k-fold cross-validation technique to evaluate the performance of different machine learning models. Furthermore, for hyperparameter tuning, we ran an inner cross-validation with in the overall cross-validation. This research methodology, depicted in Figure\ref{fig:pipeline}, allowed us to assess both the immediate and optimised effects of machine learning applications on our data sets.

\input{figures/pipeline}

\section{Data and Methods}
\subsection{Data description}

The metabolomics datasets used in this study come from a clinical study of $224$ patients scheduled for cataract surgery conducted from August $2020$ to January $2022$ at Ifocus Eye Clinic in Haugesund, Norway \cite{nilsen2023significance, jensen2024preservative, gundersen2024method,graae2023prevalence,nilsen2024significance}. The patients were sub-grouped into dry versus normal eyes based on clinical examination in a standardised manner. Of the $224$ participants, tear samples using Schirmer strips from $54$ of the dry eye positive (DED+) group and $27$ from the dry eye negative (DED-) group were selected randomly and subjected to metabolomics analysis based on a well-validated global Liquid Chromatography–Mass Spectrometry (LC-MS) method at Oslo universitetssykehus \cite{allwood2010introduction}. Comprehensive metabolome coverage was obtained by analyzing the samples in both positive and negative ionization modes on the metabolomic instrument: Electrospray Ionization Positive (ESI+) and Electrospray Ionization Negative (ESI-) \cite{hua2001comparison}. As most participants in the clinical study had dry eye disease, the number of samples with dry eye is greater than those without dry eye.
In the ESI+ mode, a total of $1922$ metabolites were identified, of which $611$ are known molecules, while the remainder are yet to be identified. Similarly, the ESI- mode revealed $939$ metabolites, with $401$ known and the rest currently unidentified. The metabolomics study followed the tenets of the Declaration of Helsinki and was approved by the Regional Committee for Medical and Health Research Ethics in Norway (Reference number $2020$/$140664$). The datasets were anonymized prior to the machine learning experiments and examined group-wise, rendering no information that could be traced back to individuals in the clinical study. Hence, the regional ethics committee was requested, and due to the true anonymity of the datasets, additional approval was not needed.

The metabolomics data sets from both ionization modes were merged, aiming to utilize all features of the patients for feeding the machine learning model. This merged data set resulted in a total of $2861$ metabolites, out of which $1012$ are known and the remaining $1849$ are yet unidentified. The integration of metabolomics data sets from both ionization modes was designed to augment the data set and enhance the predictive capability of machine learning models for dry eye disease in cataract patients. Information about the metabolomics data sets, categorized by each ionization mode and the merged overview, is provided in Table \ref{tab:data_info}.

\input{tables/data_info}

\subsection{Data preprocessing} 

The data sets were checked for missing values using a snippet of Python code automatically before normalization, and no missing values were found. Normalization is important in metabolomics data sets to mitigate the effects of variation arising from instrument sensitivity \cite{sun2023pretreating,misra2020data}. To address disparities in high and low-intensity features and to decrease the variability in data spread (heteroscedasticity), we standardized each metabolite's logarithmic value. This standardization process consisted of subtracting the mean $\bar{x}$ of the logarithm of each metabolite value and dividing by its standard deviation (s) \cite{van2006centering}. Next, quantile normalization was applied with the aim of further minimizing sample-to-sample variation, thus ensuring a more uniform data structure to subsequent analyses \cite{jauhiainen2014normalization,alakwaa2018deep,meena2022application,li2016performance}.

\input{equations/quant}

\subsection{Machine learning models}

For classifying cataract patients based on dry eye disease status, six supervised machine learning techniques were employed, and two dummy classifiers with different strategies were used as baselines for the performance benchmark. These particular supervised machine learning models were chosen because they represent a range of common methods in metabolomics research \cite{galal2022applications}. Furthermore, the machine learning models such as tree-based models demonstrate that they are more appropriate methods for tabular data sets than deep learning models and also require much less tuning and computational resources \cite{shwartz2022tabular,grinsztajn2022tree}. Additionally, this study has limited data available, which is generally not sufficient for deep learning models. The methods that were employed in this study are as follows:

\gls{xgboost}: \gls{xgboost} \cite{chen2016xgboost} operates as an ensemble learning method, utilizing a foundation of decision tree models. Designed on the concept of decision trees, it employs a gradient boosting (GB) system and is a modular method of tree boosting commonly used in machine learning. XGBoost integrates collections of classifiers with initially low precision through a process of iterative refinement, culminating in a classifier of superior accuracy. It offers parallel tree boosting, also known as \gls{gbdt} or \gls{gbm}, to address a range of data science problems quickly and accurately. Given its proficiency in managing high-dimensional sparse data, and ability to evaluate the importance of features, XGBoost stands out as a particularly suitable choice for analysing metabolomics data sets. Additionally, XGBoost has been widely used in various real-world applications, such as predicting disease outcomes, due to its computational efficiency and scalability \cite{yuan2024discrimination}.

\gls{rf} \cite{breiman2001random}: \gls{rf} is a type of classification model that creates numerous decision trees, each developed from various subsets of randomly chosen input variables \cite{cutler2007random}. This method is resistant to overfitting due to its ensemble nature and does not require scaling prior to analysis. In essence, \gls{rf} is an effective classification tool that offers an inherent unbiased estimate of the error in generalisation, and is also resistant to anomalies \cite{strobl2009introduction}.

\gls{svm} \cite{cortes1995support}: \gls{svm} represent a classification approach that differs from conventional methods by focussing not just on minimising training errors, but on reducing the upper limit of the generalisation error by maximising the margin between the training data and the separating hyperplane. An advantage of using SVMs, particularly with the Radial Basis Function (RBF) kernel, is their ability to handle non-linear relationships effectively. The RBF kernel transforms the input space into a higher-dimensional space where the data points can be linearly separable, allowing the creation of a more flexible decision boundary \cite{cervantes2020comprehensive}.

\gls{mlp} \cite{baum1988capabilities}: \gls{mlp} is a variant of feedforward \gls{anns}. It is structured around three principal layers: an input layer, one or more hidden layers, and an output layer. \gls{mlp}s operate by forwarding input through the network in a single direction, through layers of unidirectionally connected nodes, and are commonly trained using the backpropagation technique. They incorporate non-linear activation functions, which transform the incoming signal at a node into the outgoing signal, facilitating the network's ability to model complex relationships.

\gls{lr} \cite{hosmer2013applied}: Logistic Regression is a statistical approach designed to examine binary outcomes by establishing the connection between the outcome and various predictor variables. This technique is widely applied in fields such as medicine, biology, and epidemiology to solve binary classification challenges, where the goal is to categorise data into one of two groups \cite{nusinovici2020logistic}. When dealing with data sets that contain a vast array of features, there's a risk that logistic models might overfit. To address this, regularization methods come into play. This study employs L2 regularization (ridge) \cite{ng2004feature} within its algorithm, which is why we describe our method as logistic ridge regression. L2 regularization helps reduce overfitting by adding a penalty term proportional to the square of the feature weights. It prevents the model from overweighting certain features, resulting in better generalization \cite{lewkowycz2020training}.

\gls{k-nn} \cite{franco2001estimation}: \gls{k-nn} algorithm is a non-parametric method used for classification and regression, which predicts the label of a sample by aggregating the labels of its k nearest neighbors in the feature space. Its benefits include simplicity, effectiveness in handling multi-class cases, and the ability to adapt its model based on the local data structure.

\gls{dc}: \gls{dc} generates outcomes without considering the input features. It acts as a baseline for comparing the effectiveness of more sophisticated classification models and is not intended for practical problem-solving applications. In this study, the 'uniform' and 'most frequent' strategies were selected for the \gls{dc}, which we refer to as DCU and DCM, respectively. The 'uniform' strategy predicts each class with equal probability, regardless of its frequency in the data set, thereby serving as a baseline to ensure that any predictive model performs better than chance. The 'most frequent' strategy always predicts the most common class observed in the training data set and serves as a baseline for the highest possible accuracy that any model can achieve by simply predicting the most frequent class.

\subsection{Evaluation metrics}

To accurately assess the effectiveness of the eight machine learning models deployed, it is important to employ suitable evaluation metrics. Consequently, in this study, a set of criteria was utilized for the evaluation, including the \gls{auc}, Balanced Accuracy, \gls{mcc}, Specificity, and the F1-Score. The selected metrics provide a clear view of the performance of the machine learning models when applied to metabolomics data sets, ensuring a reliable analysis of their predictive capabilities and overall accuracy.

The \gls{auc} in binary classification refers to the area under the \gls{roc} curve. This curve plots the true positive rate against the false positive rate at various threshold settings, providing a measure of a model's ability to discriminate between the positive and negative classes across all possible thresholds. In this context, an \gls{auc} value close to $1.0$ indicates perfect prediction, while an \gls{auc} value of $0.5$ suggests performance no better than chance. A higher \gls{auc} value indicates a stronger capability of the model to distinguish between cataract patients with and without dry eye. The \gls{auc} is often considered a reliable performance metric for imbalanced binary classification problems \cite{liu2008exploratory}. However, when the data set is imbalanced and the \gls{auc} has reached a high score, the classification performance may not be as perfect as the \gls{auc} value suggests. Therefore, we used other metrics to reflect different aspects of machine learning capability and compared them in the metabolomics data sets.

In the following metrics, TP represents the count of samples with dry eye disease correctly identified by the model. TN denotes the count of samples without dry eye disease accurately predicted as negative. In contrast, FP refers to samples without dry eye disease that were incorrectly predicted to have the disease, while FN accounts for samples with dry eye disease that were mistakenly classified as negative by the model.

The specificity metric indicates the proportion of correctly identified negative instances. It highlights a model's proficiency in accurately classifying negative classes. This metric allows us to understand the model's ability to identify cataract patients who don't have dry eye, which is important given the lower prevalence of the negative class in the data sets of this study. The formula for specificity is provided in \eqref{speci}.

\input{equations/spec}

The sensitivity assesses the percentage of positive instances that are accurately identified. It addresses the issue of how many actual positives are properly classified. In this study, the sensitivity metric provides insight into the model's effectiveness in correctly detecting dry eye, which is the majority class. The formula for sensitivity is in \eqref{sens}.

\input{equations/sens}

Balanced accuracy \cite{jiao2016performance} calculates the average of specificity and sensitivity, making it a valuable metric for scenarios where there are unequal sample sizes across classes. It is appropriate for this study since the data sets are imbalanced, as it helps prevent overestimation of performance evaluations. This metric assesses the accuracy of each class individually and then determines the simple average of these accuracies, as demonstrated in \eqref{bal_acc}.

\input{equations/bal_acc}

The \gls{mcc} metric is appropriate for evaluating binary classification models, particularly when faced with imbalanced distributions of classes. By including true positives, true negatives, false positives, and false negatives in its calculation, \gls{mcc} offers a balanced evaluation of a model's accuracy. This metric provides insight into the model's true predictive power in the context of predicting dry eye disease in cataract patients, where classes are imbalanced. The \gls{mcc} formula is demonstrated in \eqref{mcc}.

\input{equations/mcc}

The F1-score is a harmonic mean of precision and sensitivity. This balance ensures that both sensitivity and precision are taken into account which discourages extreme values. In this study, the F1-score metric provides a balanced view of the model's performance by correctly identifying cataract patients with dry eye (sensitivity) while minimizing the misclassification of cataract patients without dry eye (precision). The calculation of the F1-score follows \eqref{f1}.

\input{equations/f1}

\subsection{Code and Availability}

The experiments for this study were conducted using the Python programming language within Google Colaboratory. This environment provided a \gls{cpu} backend, equipped with $13$ GB of RAM and $108$ GB of storage space. The machine learning libraries utilized in the research included sklearn version $1.2.2$, numpy version $1.23.5$, seaborn version $0.13.1$, pandas version $1.5.3$, matplotlib version $3.7.1$, and scipy version $1.11.4$.
The Jupyter notebooks employed in this research are accessible at the following location: \url{https://github.com/sajadamouei/classification-metabolomics}

\section{Experiments and Results}

To explore the predictive capacity of machine learning models in imbalanced metabolomics binary classification data sets, we adopted an evaluation methodology. Data sets reflect a high-dimensional data structure common to metabolomics studies. Given the challenges posed by the data set's imbalance and complexity, we implemented an approach to model development and validation to ensure the reliability of our findings.

Initially, we developed eight machine learning models on the entire data set. These models were selected based on their ability to handle challenges in the data set and their various learning mechanisms, providing a broad range of evaluations. The development and evaluation of these models were carried out using a stratified 10-fold cross-validation technique. This technique entails dividing the data set into ten folds of equal size, ensuring that the distribution of labels is consistent across all folds to maintain the representativeness of the data set. In each iteration, nine folds were used to train the model, and the remaining fold served as a test set. This process was repeated ten times, each fold serving as a test set once, ensuring that every data point was used for both training and testing. The performance of each model was then evaluated based on the mean of the metrics obtained from the ten iterations, providing an aggregate measure of the model's effectiveness across the entire data set.

In the field of machine learning, hyperparameters are variables that define the model's structure or the characteristics of the learning algorithm, and are set before training begins. Unlike hyperparameters, other parameters like coefficients or weights change during training. The need and number of hyperparameters differ depending on the algorithm. To examine the impact of the model optimization, we enhanced our evaluation methodology by adopting a nested stratified cross-validation strategy for fine-tuning these hyperparameters.

Given the specific characteristics and challenges of our data sets, we opted for a limited search space tailored to each model to prevent overfitting, utilizing a grid search method for this purpose. In this approach, an additional stratified 5-fold cross-validation dedicated to hyperparameter tuning was conducted within each fold of the primary stratified 10-fold cross-validation. This nested setup enabled us to further partition the training folds from the outer loop into smaller sub-folds. Here, four sub-folds were employed for training with different hyperparameter settings, while the fifth sub-fold acted as a validation set to evaluate these settings. The best-performing hyperparameter set on the validation sets from the five inner folds was then selected as the optimal configuration for each model. With these optimal hyperparameters, the model was trained anew on the complete training data set from the outer loop, prior to its assessment on the test fold.

Algorithm \ref{cv_alg} illustrates the process of stratified 10-fold cross-validation, including the incorporation of nested stratified 5-fold cross-validation for hyperparameter tuning.

\input{equations/cv_algo}

\subsection{Results of Hyperparameter tuning}

To improve the performance of machine learning models, optimization and hyperparameter tuning for models such as \gls{xgboost}, \gls{rf}, \gls{svm}, \gls{mlp}, \gls{lr}, and \gls{k-nn} were performed using the grid search in a nested stratified k-fold cross-validation approach. \gls{auc} metrics were used in the grid search to obtain the optimal hyperparameters for the machine learning models. In this study, the \gls{lr} model was selected as the most suitable based on its performance across all evaluation metrics, which will be presented in detail in the next subsection. For the hyperparameters of the \gls{lr}, the L2 penalty was chosen as the regularization method to help prevent overfitting. The 'max\_iter' hyperparameter was set to $100$ to define the maximum number of iterations allowed for the solvers to converge. Other hyperparameters selected for tuning included 'C', which controls the inverse of the regularization strength; 'solver', the algorithm used to optimize the model parameters; and 'class\_weight', which addresses imbalances in the classes. These hyperparameters were tuned using the nested k-fold cross-validation approach and a grid search strategy. The search space and the optimal hyperparameters identified for the \gls{lr} model are presented in Table \ref{table:lr_hype}.

\input{tables/hyp_tuning}

\subsection{Performance of machine learning models}

Tables \ref{tab:res_esi+} to \ref{tab:res_merge} summarize the efficacy of different machine learning models across three metabolomics data sets, using five evaluation metrics. The values presented in these tables are the averages obtained from the 10-fold cross-validation process. First, we examine performance obtained by machine learning models in three metabolomics data sets based on each metric, so that we can identify which data set is most suitable for machine learning models to detect dry eye in cataract patients.

The best \gls{auc} result achieved by machine learning models on the merged data set was $0.8378$, surpassing its effectiveness on the ESI- and ESI+ data sets by $0.0495$ and $0.0211$, respectively. The highest balanced accuracy recorded by machine learning models was $0.735$ in both the merged and ESI+ data sets, outperforming the ESI- data set by $0.0233$. Similarly, for the \gls{mcc} metric, models performed better on the merged and ESI+ data sets, achieving a score of $0.5147$, which is an improvement of $0.0801$ over the ESI- data set. In terms of specificity, models scored highest on the ESI- data set at $0.7167$, exceeding scores in the merged and ESI+ data sets by $0.15$. For the F1-Score, the peak performance was observed in the ESI+ data set at $0.8669$, higher than in the merged and ESI- data sets by $0.0092$ and $0.0174$.

These results suggest that models generally performed better on the merged data set in terms of \gls{auc}, balanced accuracy, and \gls{mcc}, though they fell slightly short in specificity and F1-Score. Therefore, these outcomes indicate that the merged data set can enhance model performance across some metrics, and achieve more balanced performance across all metrics, suggesting a beneficial effect of merging the ESI+ and ESI- data sets on model efficacy.

Following, we explore the effectiveness of various machine learning models, focusing on the merged data set, as detailed in Table \ref{tab:res_merge}, to find the most suitable models for identifying dry eye in cataract patients. The results in Table \ref{tab:res_merge} reveal that, compared to baseline approaches that involve random guesses or selecting the most frequent class (represented by Dummy Classifiers), almost all the models demonstrated better performance across multiple evaluation metrics.

According to Table \ref{tab:res_merge}, the \gls{lr} model has the highest \gls{auc} at $0.8378$. The \gls{auc} metric, commonly used in medical machine learning applications for disease prediction problems, shows that the \gls{lr} model effectively differentiates between the two classes. Compared to other tested machine learning methods, this model achieves a higher \gls{auc}. After \gls{lr}, the \gls{rf} and \gls{xgboost} models recorded the next highest scores of $0.7878$ and $0.7861$, respectively. Due to the imbalance in the data set's classes, other metrics besides the \gls{auc} were also considered to provide a clearer comparison of the models.

In terms of balanced accuracy, \gls{lr} achieved the top score of $0.735$, outperforming other models in the merged data set. This metric highlights the model's effectiveness on an imbalanced data set by valuing the performance across both minority and majority classes equally. \gls{xgboost} and \gls{rf} followed with scores of $0.72$ and $0.675$, respectively.

For the \gls{mcc}, \gls{lr} again led with a score of $0.5147$, indicating its better performance in handling imbalanced data sets compared to other models. \gls{xgboost} and \gls{rf} followed with scores of $0.4639$ and $0.42$, respectively.

For specificity, the \gls{lr} and \gls{xgboost} achieved the highest score of $0.5667$, and \gls{mlp} followed with a score of $0.45$. The negative samples are the minority classes in the data sets, making this metric important for demonstrating the models' ability to correctly detect negative samples.

Regarding the F1-score, the \gls{rf} model outperformed the others with a score of $0.8577$, closely followed by \gls{lr} and \gls{xgboost} with scores of $0.8513$ and $0.8337$, respectively.

\input{tables/results}

In summary, this comparative analysis indicates that the \gls{lr} model achieved the highest performance in \gls{auc}, balanced accuracy, \gls{mcc}, and specificity compared to other machine learning models in the merged data set. With a performance close to highest score in F1-score, it maintained balanced performance across all metrics compared to other machine learning models in this study on the merged metabolomics data set.

Previous research shows that the \gls{lr} model was successful in biological and clinical contexts \cite{waddington2020using, boateng2019review}. The results of this study demonstrated that the \gls{lr} model outperforms other machine learning models and suggests that it is possible to achieve high performance even with a quite simple model. The more complex models, can quickly suffer from overfitting on the fairly small data set in this study. Additionally, \gls{lr} is generally computationally less intensive and achieves convergence faster than more complex methods. This could be another reason for more stable model training and prediction, especially in this study's data set, characterised by high-dimensional spaces and a limited number of samples. This is further discussed in Section \ref{sec:cons}.

The next best machine learning models following the \gls{lr} model on the merged data set are \gls{xgboost} and \gls{rf}, which are ensemble models. They achieved satisfactory performance on some evaluation metrics, but underperformed on others, and overall did not achieve consistent performance across all evaluation metrics.

In this study, the weakest results were observed with \gls{k-nn} and \gls{svm}. Despite the theoretical benefits often associated with \gls{svm}, especially with an RBF kernel, in the field of bioinformatics and for high-dimensional data sets \cite{galal2022applications, zheng2017predictive, corona2018svm}, the results presented in Table \ref{tab:res_merge} demonstrate performance comparable to that of baseline dummy classifiers across several evaluation metrics on the merged data set. Both the \gls{k-nn} and \gls{svm} models demonstrated suboptimal outcomes in this study. These results emphasize the need for a careful and nuanced approach when selecting models for metabolomics data sets.

Furthermore, integrating data sets from different ionization modes into a single data set has proven to improve model training by providing a wider range of features and patterns. This method enhances the data pool and helps models generalize better and increase prediction accuracy.

\subsection{Model Performance Consistency}
\label{sec:cons}

Figure \ref{fig:auc_sd} shows classification performance in terms of \gls{auc} along the y-axis and the standard deviation of the measured \gls{auc} from the different cross-validation folds along the x-axis. The standard deviation therefore gives an indication of the level of consistency in performance for different machine learning methods under repeated training of the algorithm. If a method has high classification performance over the folds, we also expect high classification constancy over the folds. Or said in another way, it is not possible to achieve high classification performance and at the same time have low consistency. Inspecting the three panels in Figure \ref{fig:auc_sd}, we see that this is the overall trend, but there are for sure also differences in consistency for methods with about the same classification performance. Intuitively, we can expect that algorithms that are simple to train with a convex loss landscape, such as logistic ridge regression (based on a linear classifier), might document higher constancy compared to models that have a more complex loss landscape. We see that logistic ridge regression documents high consistency relative to classification performance for all the three data sets (panels (\ref{fig:sub_auc_sd_esi+}) $-$ (\ref{fig:sub_auc_sd_merge})). 

Among the methods with a mean \gls{auc} over $0.75$ for the ESI+ data set (panel (\ref{fig:sub_auc_sd_esi+})), we see that, in addition to logistic ridge regression, \gls{xgboost} also documents high constancy, while \gls{mlp}, \gls{svm}, and \gls{rf} document less consistency. Further, \gls{k-nn} also documents poor constancy, which is as expected since the mean \gls{auc} is low. 

Among the methods with a mean \gls{auc} over $0.7$ for the ESI- data set (panel (\ref{fig:sub_auc_sd_esi-})), we see that only logistic ridge regression documents high consistency, while \gls{mlp} and \gls{xgboost} document medium consistency. The other methods document poorer classification performance, and as expected, consistency is also low.

Among the methods with a mean \gls{auc} over $0.75$ for the merged data (panel (\ref{fig:sub_auc_sd_merge})), we see that in addition to logistic ridge regression, \gls{xgboost} also documents high consistency, while \gls{mlp}, \gls{rf}, and \gls{svm} document medium consistency.

To summarize, logistic ridge regression documents high constancy, \gls{xgboost} medium to high consistency in classification performance while the other methods document medium to low consistency.

\input{figures/auc_sd}

\subsection{Logistic Regression Model Interpretation}

Table \ref{tab:features} displays the features with the highest values of the estimated regression parameters, which refer to the molecules most strongly associated with dry eye. All the molecules among the ten highest values, except one, including Militarinone A, unknown303, trans-Anethole, unknown340, unknown1175, hydroxymethyl, Hypoxanthin, unknown1265, and unknown439, have positive estimates indicating their association with an increased risk of dry eye. One molecule, unknown343, has a negative estimate, suggesting its association with a reduced risk of dry eye. This was expected, as the data sets are imbalanced, with a majority being cataract patients with dry eye disease.
\input{tables/feat_impo}

Table \ref{tab:features_known} displays the known molecules with the highest regression parameters. The known molecules not included in Table \ref{tab:features} are about equally divided between positive and negative associations with dry eye disease. We further observe that different known molecules are included from the ESI+ and ESI- collection.

\input{tables/feat_impo_known}

\section{Conclusion}

In this study, we focused on applying machine learning techniques to classify cataract patients, distinguishing between those with and without dry eye disease in a cohort of patients scheduled for cataract surgery. Dry eyes adversely affect patients' quality of life and represent a risk factor for patients undergoing cataract surgery. Selecting the most suitable machine learning models for metabolomics data sets is crucial because no single model excels in all scenarios, and their accuracy and interpretability can influence subsequent steps in the application of these models in the metabolomics field. Therefore, in response to the challenge of selecting the optimal machine learning model for our specific imbalanced binary classification metabolomics data sets, we evaluated eight machine learning models to determine the most effective method for classifying cataract patients with and without dry eye disease.

To assess these models, we employed k-fold cross-validation across the entire data sets and used nested k-fold cross-validation for tuning hyperparameters. Additionally, multiple evaluation metrics were utilized to gain a clearer understanding of the models' behavior and effectiveness. This is particularly important because the data sets in this study were imbalanced.

The results highlighted that the \gls{lr} model outperformed other machine learning models on the merged metabolomics data set, achieving an \gls{auc} of $0.8378$, balanced accuracy of $0.735$, and \gls{mcc} of 0$.5147$. It also achieved a specificity of $0.5667$ and an F1-score close to that of the best model, with values of $0.8513$, demonstrating overall balanced performance across all evaluation metrics. The \gls{lr} model indicates consistent performance across 10-fold cross-validation, making it a reliable machine learning model in this study compared to other models in the metabolomics data set. \gls{xgboost} and \gls{rf} showed good performance on some evaluation metrics but could not maintain consistent performance across all metrics and 10-folds. \gls{svm} with RBF kernel and \gls{k-nn} methods also showed poor performance in this study.

Furthermore, the results of this study demonstrate that the integration of data sets with positive and negative ionization modes enriches the models' training process. This contributes to a more nuanced feature set that enhances generalization capabilities and prediction accuracy, positioning data set merging as a strategic approach to improve machine learning model efficacy for these specific metabolomics data sets.

The \gls{lr} model outperformed other models in this study, and the results obtained were promising. We believe that the \gls{lr} model could aid medical experts in detection of dry eye disease in cataract patients with metabolomics data sets.

\section{future works}

Explainable artificial intelligence (XAI) methods can be explored with the most effective machine learning models to interpret the best performing models and identify metabolites that are potentially associated with dry eyes for cataract patients. 


\end{document}

%% file: glossary.tex
\newacronym{mcc}{MCC}{Matthews Correlation Coefficient}
\newacronym{svm}{SVM}{Support Vector Machine}
\newacronym{xgboost}{XGBoost}{Extreme Gradient Boosting}
\newacronym{mlp}{MLP}{Multilayer Perceptron}
\newacronym{sgd}{SGD}{Stochastic Gradient Descent classifier}
\newacronym{k-nn}{k-NN}{K-Nearest Neighbors}
\newacronym{cpu}{CPU}{Central Processing Unit}
\newacronym{auc}{AUC}{Area Under the Curve}
\newacronym{roc}{ROC}{Receiver Operating Characteristic}
\newacronym{anns}{ANNs}{Artificial Neural Networks}
\newacronym{ded}{DED}{Dry Eye Disease}
\newacronym{lr}{LR}{Logistic ridge Regression}
\newacronym{rf}{RF}{Random Forest}
\newacronym{dc}{DC}{Dummy Classifier}
\newacronym{gbdt}{GBDT}{Gradient Boosting Decision Tree}
\newacronym{gbm}{GBM}{Gradient Boosting Machine}

%% file: figures/pipeline.tex
\begin{figure*}[htbp]
    \centering
    \includegraphics[width=\textwidth]{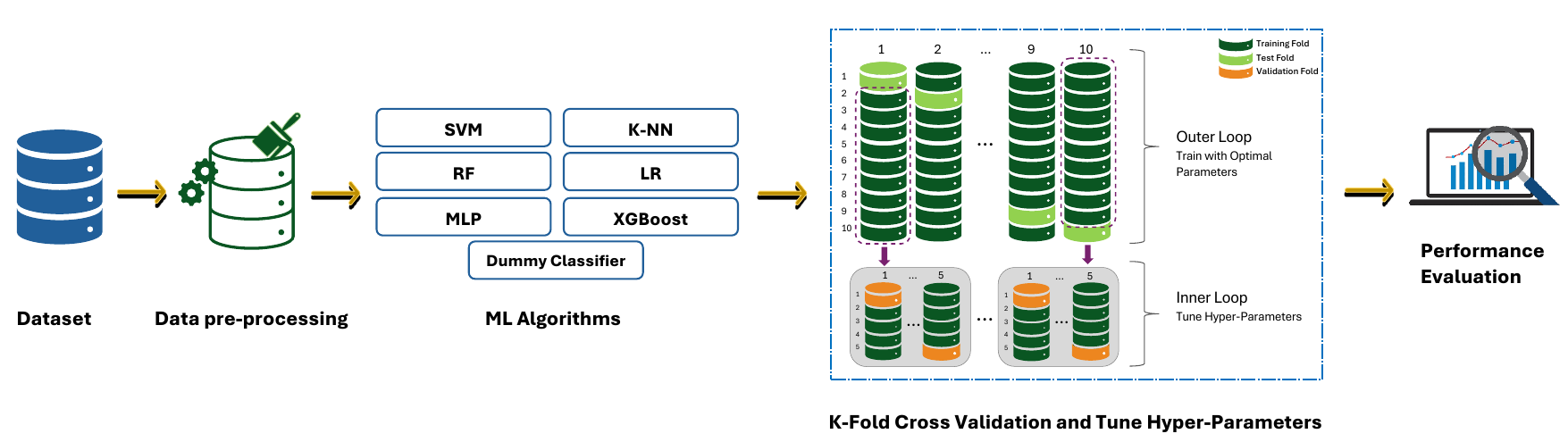}
    \caption{Operational Flow of proposed Dry Eye Disease classification.}
    \label{fig:pipeline}
\end{figure*}

%% file: tables/data_info.tex
\begin{table}[htbp]
    \centering
    \captionsetup{font=scriptsize}
    \caption{Metabolomics datasets information}
    \begin{tabular}{l c c c}
        \toprule
         & M1 ESI+ & M1 ESI- &  Merged ESI+ \& ESI- \\
         \midrule
         Samples & $81$ & $81$ & $81$\\
         Known Metabo. & $611$ & $401$ & $1012$  \\
         Unknown Metabo. & $1311$ & $538$ & $1849$  \\
         All metabo. & $1922$ & $939$ & $2861$ \\
         class 0 & $27$ & $27$ & $27$ \\
         class 1 & $54$ & $54$ & $54$ \\
         \bottomrule
    \end{tabular}
    \begin{tabular}{p{0.7\linewidth}}
        Abbreviations: Metabo. = Metabolites
    \end{tabular}
    \label{tab:data_info}
\vspace{8pt}
\end{table}

%% file: equations/quant.tex
\begin{equation}
    \hat{x}_{i j}=\left(\frac{\log _2\left(x_{i j}\right)-\overline{x_i}}{s}\right)\label{quant}
\end{equation}

%% file: equations/spec.tex
{
\scriptsize
\begin{equation}
    \text{Specificity} = \frac{TN}{TN + FP}\label{speci}
\end{equation}

}

%% file: equations/sens.tex
{
\scriptsize
\begin{equation}
    \text{Sensitivity} = \frac{\text{TP}}{\text{TP} + \text{FN}}\label{sens}
\end{equation}
}

%% file: equations/bal_acc.tex
{
\scriptsize
\begin{equation}
    \text{Balanced Accuracy} = \frac{1}{2} \left( \frac{TP}{TP + FN} + \frac{TN}{TN + FP} \right)\label{bal_acc}
\end{equation}
}

%% file: equations/mcc.tex
{
\scriptsize
\begin{equation}
    MCC = \frac{TP \times TN - FP \times FN}{\sqrt{(TP+FP)(TP+FN)(TN+FP)(TN+FN)}}\label{mcc}
\end{equation}
}

%% file: equations/f1.tex
{
\scriptsize
\begin{equation}
    F1 = 2 \times \frac{\text{precision} \times \text{sensitivity}}{\text{precision} + \text{sensitivity}}\label{f1}
\end{equation}

}

%% file: equations/cv_algo.tex
\begin{footnotesize}
\begin{algorithm}
\caption{Stratified 10-Fold Cross-Validation}\label{cv_alg}
\begin{algorithmic}
\STATE \textbf{Input:} Dataset $D$ with $N$ instances, each comprising a feature set and a class label.
\STATE \textbf{Output:} Mean performance metrics of the models.
\STATE \textbf{Procedure:}
\begin{enumerate}
    \item Stratify $D$ by class labels to ensure class distribution is mirrored across folds.
    \item Divide $D$ into 10 equal parts, $F_1, F_2, \ldots, F_{10}$, maintaining stratification.
    \item For each fold $F_i$:
    \begin{enumerate}
        \item Treat $F_i$ as the test set, and the remaining folds as the training set.
        \item If tuning is needed, perform a stratified 5-fold CV on the training set to optimize hyperparameters.
        \item Train the model on the training set, using optimized parameters if applicable.
        \item Predict the true label for each data point in the test set, $F_i$ and store the predictions.
    \end{enumerate}
    \item Evalute the performance by comparing the the predictions from all the test sets $F_1, \ldots, F_{10}$ with true labels.
\end{enumerate}
\end{algorithmic}
\end{algorithm}
\end{footnotesize}

%% file: tables/hyp_tuning.tex
\begin{table}[ht]
\centering
\captionsetup{font=scriptsize}
\begin{tabular}{lcc}
\hline
Hyperparameter & Search Area & Optimal Value \\
\hline
C & 1.0, 0.1 & 1.0 \\
Solver & \texttt{lbfgs}, \texttt{newton-cg} & \texttt{lbfgs} \\
Class Weight & None, \texttt{balanced} & Both \\
\hline
\end{tabular}
\caption{Hyperparameter Tuning for Logistic Regression Model}
\label{table:lr_hype}
\end{table}

%% file: tables/results.tex
\begin{table}[htbp]
    \centering
    \vspace{8pt}
    \captionsetup{font=scriptsize}
    \caption{Comparative Performance of Machine Learning Models on the ESI+ data set. The values presented are the averages obtained from the 10-fold cross-validation process.}
    \begin{tabular}{l c c c c c}
        \toprule
        Model & AUC & B. A. & MCC & Spec. & F1-Score \\
         \midrule
         XGB & \textbf{0.8167} & 0.6667 & 0.3345 & 0.5167 & 0.7937 \\
         RF & 0.7631 & 0.7083 & 0.4776 & 0.4333 & \textbf{0.8669} \\
         SVM & 0.7733 & 0.6517 & 0.3495 & 0.3833 & 0.8304 \\
         MLP & 0.7494 & 0.6117 & 0.2579 & 0.55 & 0.6885 \\
         LR & 0.8111 & \textbf{0.735} & \textbf{0.5147} & \textbf{0.5667} & 0.8513 \\
         K-NN & 0.6292 & 0.6217 & 0.2832 & 0.3 & 0.8236 \\
         DCM & 0.5 & 0.5 & 0 & 0 & 0.7987 \\
         DCU & 0.5 & 0.4867 & -0.0319 & 0.1 & 0.7522 \\
         \bottomrule
    \end{tabular}
    \vspace{2mm}
    \begin{tabular}{p{0.7\linewidth}}
         Abbreviations: B. A. = Balanced Accuracy; Spec. = Specificity.
    \end{tabular}
    \label{tab:res_esi+}
\vspace{8pt}

    \captionsetup{font=scriptsize}
    \caption{Comparative Performance of Machine Learning Models on the ESI- data set. The values presented are the averages obtained from the 10-fold cross-validation process.}
    \begin{tabular}{l c c c c c c}
        \toprule
        Model & AUC & B.A. & MCC & Spec. & F1-Score \\
         \midrule
         XGB & \textbf{0.7883} & \textbf{0.7117} & \textbf{0.4346} & 0.5167 & \textbf{0.8495} \\
         RF & 0.5683 & 0.5283 & 0.0717 & 0.1333 & 0.7837 \\
         SVM & 0.5372 & 0.4433 & -0.1366 & 0 & 0.7367 \\
         MLP & 0.7056 & 0.6333 & 0.2592 & \textbf{0.7167} & 0.637 \\
         LR & 0.7311 & 0.615 & 0.2135 & 0.4167 & 0.7725 \\
         K-NN & 0.4506 & 0.4533 & -0.103 & 0.15 & 0.6859 \\
         DCM & 0.5 & 0.5 & 0 & 0 & 0.7987 \\
         DCU & 0.5 & 0.4867 & -0.0319 & 0.1 & 0.7522 \\
         \bottomrule
    \end{tabular}
    \vspace{2mm}
    \begin{tabular}{p{0.7\linewidth}}
         Abbreviations: B. A. = Balanced Accuracy; Spec. = Specificity.
    \end{tabular}
    \label{tab:res_esi-}
\vspace{8pt}

    \captionsetup{font=scriptsize}
    \caption{Comparative Performance of Machine Learning Models on the merged ESI+ and ESI- data set. The values presented are the averages obtained from the 10-fold cross-validation process.}
    \begin{tabular}{l c c c c c}
        \toprule
        Model & AUC & B. A. & MCC & Spec. & F1-Score \\
         \midrule
         XGB & 0.7861 & 0.72 & 0.4639 & \textbf{0.5667} & 0.8337 \\
         RF & 0.7878 & 0.675 & 0.42 & 0.3667 & \textbf{0.8577}\\
         SVM & 0.76 & 0.61 & 0.2583 & 0.3 & 0.8162 \\
         MLP & 0.7811 & 0.6567 & 0.3419 & 0.45 & 0.8071 \\
         LR & \textbf{0.8378} & \textbf{0.735} & \textbf{0.5147} & \textbf{0.5667} & 0.8513 \\
         K-NN & 0.6614 & 0.6183 & 0.2756 & 0.3333 & 0.8083 \\
         DCM & 0.5 & 0.5 & 0 & 0 & 0.7987 \\
         DCU & 0.5 & 0.4867 & -0.0319 & 0.1 & 0.7522 \\
         \bottomrule
    \end{tabular}
    \vspace{2mm}
    \begin{tabular}{p{0.7\linewidth}}
         Abbreviations: B. A. = Balanced Accuracy; Spec. = Specificity.
    \end{tabular}
    \label{tab:res_merge}
\vspace{8pt}

\label{tab:all_res}
\end{table}

%% file: figures/auc_sd.tex
{
\scriptsize
\begin{figure}[htbp]
    \centering
    \begin{subfigure}[b]{\columnwidth}
        \includegraphics[width=\linewidth]{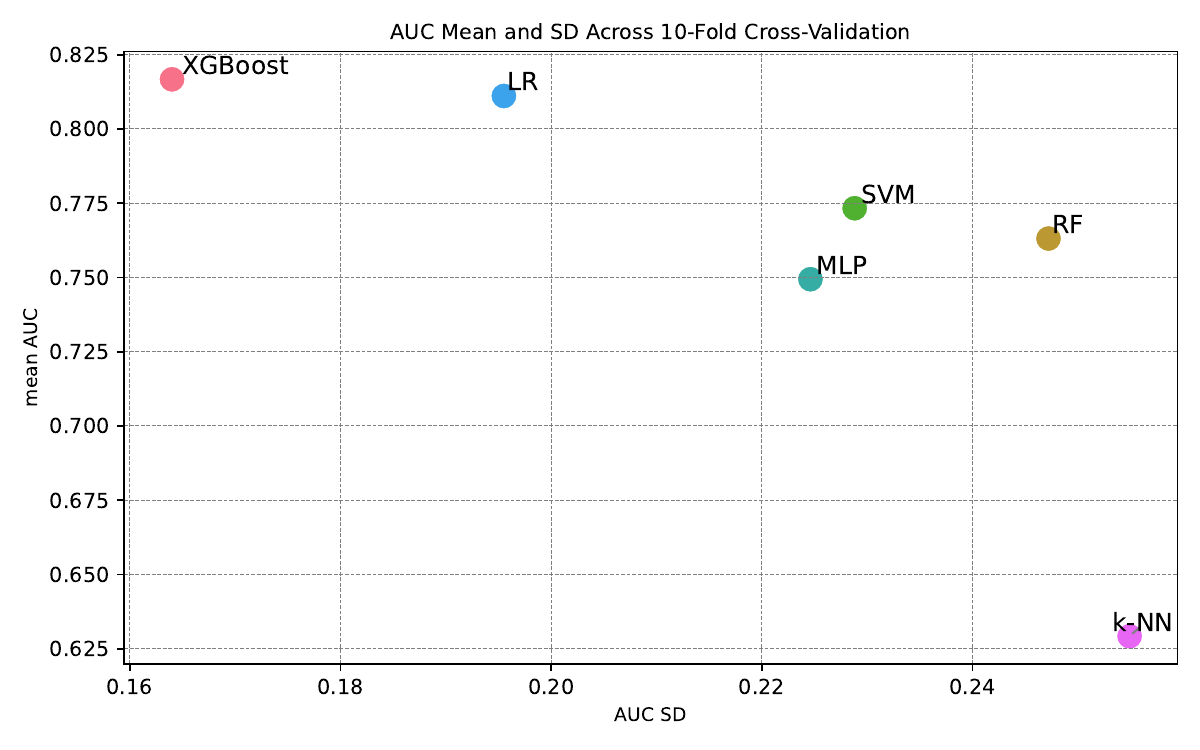}
        \caption{Model comparison on ESI+ dataset.}
        \label{fig:sub_auc_sd_esi+}
    \end{subfigure}
    
    \begin{subfigure}[b]{\columnwidth}
    \vspace{12pt}
        \includegraphics[width=\linewidth]{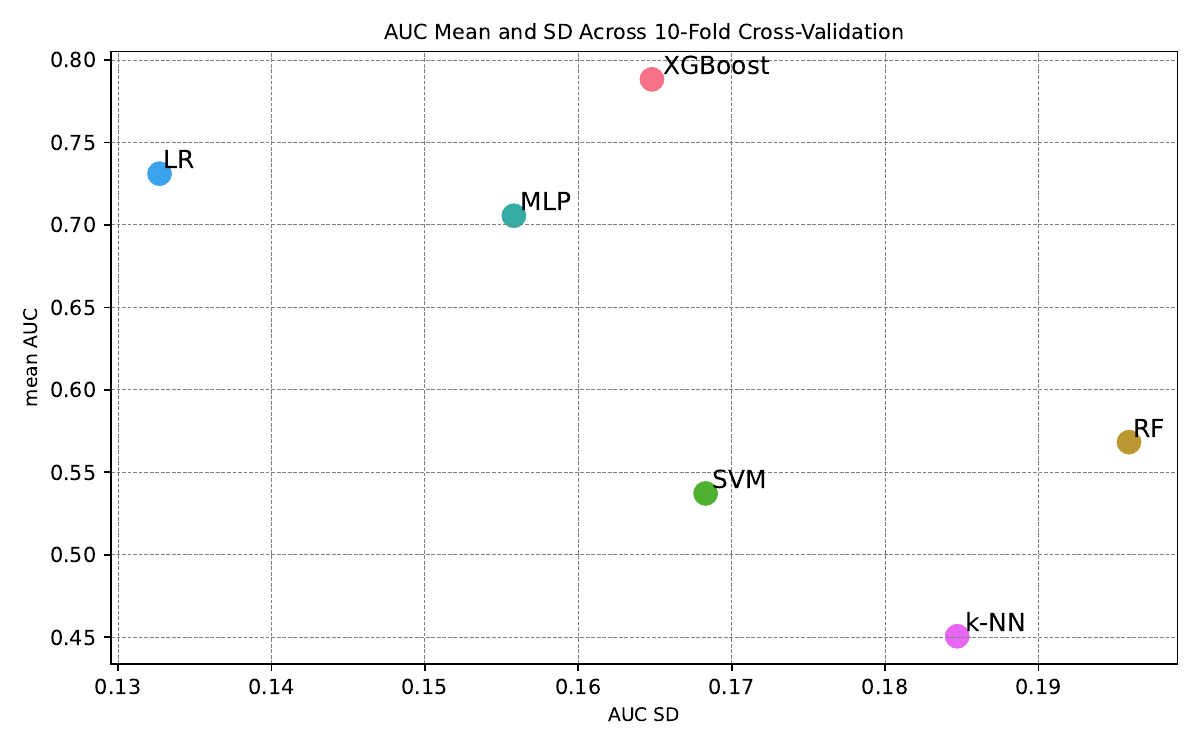}
        \caption{Model comparison on ESI- dataset.}
        \label{fig:sub_auc_sd_esi-}
    \end{subfigure}
    
    \begin{subfigure}[b]{\columnwidth}
    \vspace{12pt}
        \includegraphics[width=\linewidth]{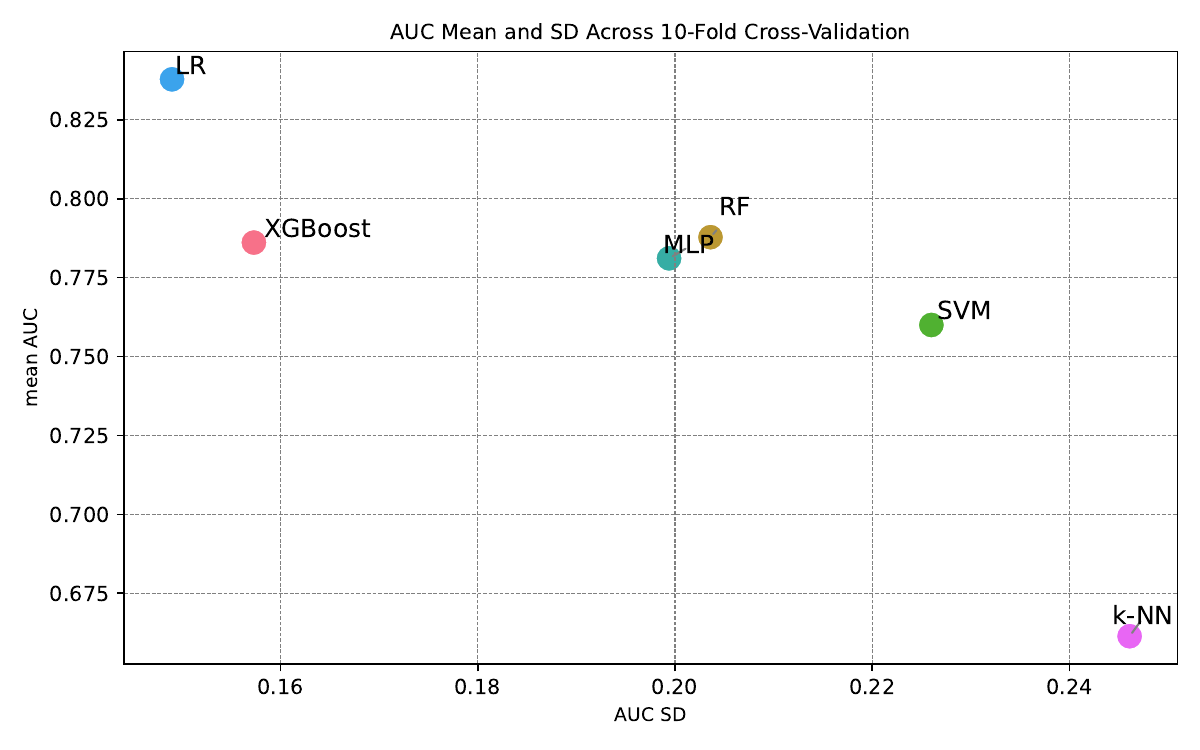}
        \caption{Model comparison on merged dataset.}
        \label{fig:sub_auc_sd_merge}
    \end{subfigure}
    
    \caption{Visualizing model performance: The graph provides a comparison of AUC scores derived from 10-fold cross-validation, highlighting the balance between mean AUC and AUC SD. The lighter and darker shades indicate base and optimized models, respectively.}
    \label{fig:auc_sd}
\end{figure}
}

%% file: tables/feat_impo.tex
\begin{table}[ht]
\centering
\captionsetup{font=scriptsize}
\begin{tabular}{cll}
\hline
\textbf{Order} & \textbf{Metabolites Name}  & \textbf{Coefficient}       \\ \hline
1              & Militarinone A\_ESI+    &  0.06868     \\
2              & unknown303\_ESI+       &   0.06539    \\
3              & trans-Anethole\_ESI+   &    0.0641    \\
4              & unknown340\_ESI+       &     0.05915   \\
5              & unknown1175\_ESI+      &     0.05755   \\
6              & hydroxymethyl\_ESI+     &      0.05497    \\
7              & Hypoxanthin\_ESI+       &   0.05428  \\
8              & unknown343\_ESI-        &   -0.0534  \\
9              & unknown1265\_ESI+      & 0.0533 \\
10             & unknown439\_ESI-       &  0.05314 \\ \hline
\end{tabular}
\caption{Top 10 Metabolites Influencing Logistic Regression Model Performance}
\label{tab:features}
\end{table}

%% file: tables/feat_impo_known.tex
\begin{table}[ht]
\centering
\captionsetup{font=scriptsize}
\begin{tabular}{cll}
\hline
\textbf{Order} & \textbf{Metabolites Name}  & \textbf{Coefficient}       \\ \hline
1              & Militarinone A\_ESI+    &  0.06868     \\
2              &    trans-Anethole\_ESI+    &    0.0641   \\
3              &  hydroxymethyl\_ESI+  &   0.05497     \\
4              &    Hypoxanthin\_ESI+    &    0.05428    \\
5              &    Indane\_ESI+   &    0.0512    \\
6              &   Lauramide\_ESI+   &      0.0508    \\
7              &    benzoic acid\_ESI-    &   -0.0499  \\
8              & isodesmosine\_ESI+        &   0.0446  \\
9              & ethanone\_ESI+      & -0.0425 \\
10             & MFCD03411993\_ESI+       &  -0.0420 \\ \hline
\end{tabular}
\caption{Top 10 Known Metabolites Influencing Logistic Regression Model Performance}
\label{tab:features_known}
\end{table}